\pdfoutput=1

\documentclass[11pt]{article}

\usepackage{ACL2023}

\usepackage{times}
\usepackage{latexsym}
\usepackage{amsmath}
\usepackage{inconsolata}
\usepackage{graphicx}
\usepackage{tabularx}
\usepackage{multirow}
\usepackage{mathtools}
\usepackage{enumitem,kantlipsum}
\usepackage[export]{adjustbox}
\usepackage{booktabs}
\usepackage{caption}
\usepackage{subcaption}
\usepackage{tikz}
\usepackage{cleveref}
\usepackage{hyperref}

\usepackage[T1]{fontenc}

\usepackage[utf8]{inputenc}

\usepackage{microtype}

\usepackage{inconsolata}

\newcommand{\example}[1]{\emph{``#1''}}

\DeclareMathOperator*{\argmax}{argmax}

\newcommand\data{\textsc{ThinkP}}
\newcommand{\mycomment}[1]{}
\newcommand{\mailto}[2]{\texttt{\href{mailto:#1}{#2}}}

\title{From Key Points to Key Point Hierarchy:\\ Structured and Expressive Opinion Summarization}

\author{Arie Cattan\textsuperscript{1}\thanks{\; Equal contribution. Work done while the first author was an intern at IBM Research.} \quad
        Lilach Eden\textsuperscript{2}\footnotemark[1] \quad 
        Yoav Kantor\textsuperscript{2} \quad 
        Roy Bar-Haim\textsuperscript{2} \\
        \textsuperscript{1}Computer Science Department, Bar Ilan University  \\
        \textsuperscript{2}IBM Research       \\ 
  {
  \mailto{arie.cattan@gmail.com}{\texttt{arie.cattan@gmail.com}} \quad 
    \{\mailto{lilache@il.ibm.com}{lilache},
    \mailto{yoavka@il.ibm.com}{yoavka},
    \mailto{roybar@il.ibm.com}{roybar}\}\texttt{@il.ibm.com}
  } \\
 }

\begin{document}

\maketitle

\begin{abstract}
\emph{Key Point Analysis (KPA)} has been recently proposed for deriving fine-grained insights from collections of textual comments. KPA extracts the main points in the data as a list of  concise sentences or phrases, termed \emph{key points}, and quantifies their prevalence. While key points are more expressive than word clouds and key phrases, making sense of a long, flat list of key points, which often express related ideas in varying levels of granularity, may still be challenging. To address this limitation of KPA, we introduce the task of organizing a given set of key points into a hierarchy, according to their specificity. Such hierarchies may be viewed as a novel type of \emph{Textual Entailment Graph}. We develop \data{}, a high quality benchmark dataset of key point hierarchies for business and product reviews, obtained by consolidating multiple annotations. We compare different methods for predicting pairwise relations between key points, and for inferring a hierarchy from these pairwise predictions. In particular, for the task of computing pairwise key point relations, we achieve significant gains over existing strong baselines by applying directional distributional similarity methods to a novel distributional representation of key points, and further boost performance via weak supervision.

\vspace{0.5em}

\centering
\mbox{\includegraphics[width=1.25em,height=1.25em]{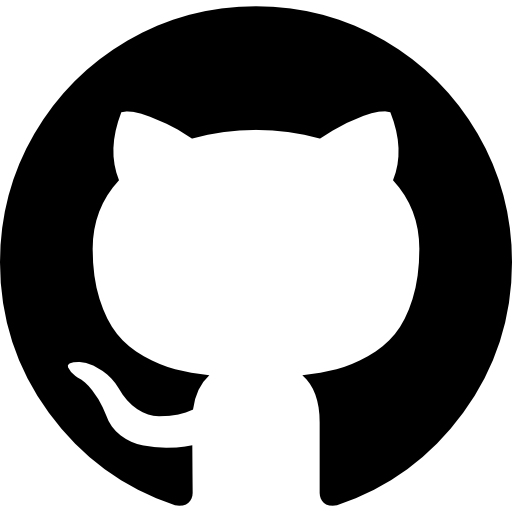}\hspace{0.5em}%
\raisebox{0.3em}{\url{https://github.com/IBM/kpa-hierarchy}}}
\end{abstract}
\section{Introduction}
Many organizations face the challenge of extracting insights from large collections of textual comments, such as user reviews, survey responses, and feedback from customers or employees. Current text analytics tools summarize such datasets via word clouds \citep{Heimerl2014WordCE} or key phrases \cite{hasan-ng-2014-automatic,AlamiMerrouni2019AutomaticKE}, which are often too crude to capture fine-grained insights. Multi-document summarization methods, on the other hand \cite{pmlr-v97-chu19b,brazinskas-etal-2020-shot,brazinskas-etal-2020-unsupervised, angelidis-etal-2021-extractive, Louis2022OpineSumES}, do not quantify the prevalence of each point in the summary, and are not well-suited for representing conflicting views \citep{bar-haim-etal-2021-every}. 
\begin{figure*}[t]
    \centering
    \includegraphics[width=\textwidth]{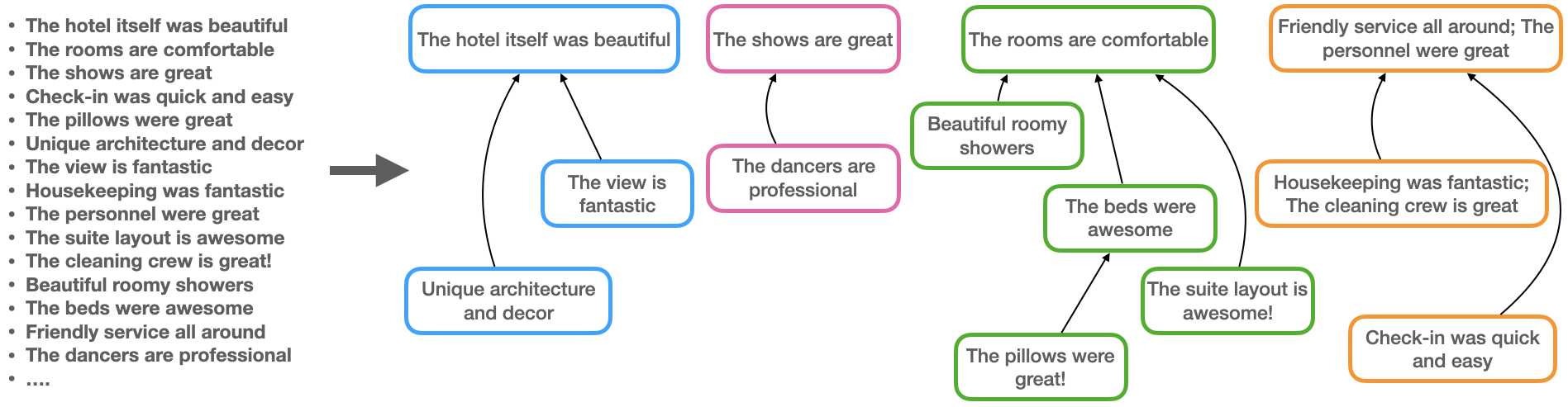}
    \caption{From a flat list of key points to a key point hierarchy (KPH). Nodes group together key points that roughly express the same idea and directed edges connect specific key points to more general ones. The number of matches for each key point is omitted.}
    \label{fig:kph_example}
\end{figure*}

\emph{Key Point Analysis (KPA)} is a recent opinion summarization framework that aims to address the above  limitations \cite{bar-haim-etal-2020-quantitative}.  KPA extracts concise sentences and phrases termed \emph{Key Points (KPs)}, which represent the most salient points in the data, and quantifies the prevalence of each KP as the number of its matching input sentences. One remaining shortcoming of KPA, however, is that it generates a flat list, which does not capture the relations between the key points. For example, consider the sample set of key points in Figure~\ref{fig:kph_example} (left), which was automatically extracted from reviews of one of the hotels in the Yelp Open Dataset\footnote{\url{https://www.yelp.com/dataset}}. The results do not provide a high level view of the main themes expressed in the reviews. It is hard to tell which key points convey similar ideas, 
and which key points support and elaborate on a more general key point. As the number of key points in the summary increases, such output becomes even harder to consume.

In this work we introduce \emph{Key Point Hierarchies (KPH)} as a novel structured representation of opinion summaries. Organizing the key points in a hierarchy, as shown in Figure~\ref{fig:kph_example} (right), allows the user to quickly grasp the high-level themes in the summary (\emph{the hotel is beautiful}, \emph{the shows are great}, \emph{comfortable rooms}, \emph{great service}), and drill down on each theme to get more fine-grained insights, e.g., from \example{The personnel were great} to \example{check-in was quick and easy}. Furthermore, key points that (nearly) convey the same meaning  (e.g., \example{Housekeeping was fantastic}, and \example{The cleaning crew is great}) are clustered together and represented as a single node in the hierarchy. This structured output makes KPA results more consumable, informative, and easier to navigate. KPH can be viewed as a new type of textual entailment graph (§\ref{sec:bg}).

We develop \data{} (\emph{\textbf{T}ree \textbf{HI}erarchy of \textbf{N}aturally-occuring \textbf{K}ey \textbf{P}oints}), the first benchmark dataset for Key Point Hierarchies, created from KPA summaries of user reviews in multiple domains (§\ref{sec:dataset}). Due to the complexity of KPH annotation,  \data{} was created by consolidating multiple annotations, to ensure its high quality.

We explore different methods for automatic KPH construction from a given set of key points (§\ref{sec:method}). Following previous work on entailment graphs (§\ref{sec:bg}), this is formulated as a two-step approach. We first compute local scores predicting the directional relation between each pair of key points. We then construct a hierarchy guided by these local pairwise predictions. 

We present novel methods and algorithmic improvements for each of the above subtasks. In particular, for the task of predicting pairwise key point relations, we achieve significant gains over existing strong baselines by applying directional distributional similarity methods to a novel distributional representation of key points, and further boost performance via weak supervision. We release the \data{} dataset to encourage further research on this challenging task.

Overall, our work contributes to several lines of research, including key point analysis, opinion summarization, entailment graphs, and distributional methods for natural language inference. Furthermore, as we demonstrate in §\ref{subsec:data_prop}, our novel \data{} dataset captures diverse types of inferences between pairs of naturally-occurring texts, making it an interesting resource for NLI research in general.

\section{Background}
\label{sec:bg}
\paragraph*{Key Point Analysis.}
\citet{bar-haim-etal-2020-arguments, bar-haim-etal-2020-quantitative} proposed \emph{Key Point Analysis (KPA)} as a summarization framework that provides both textual and quantitative summary of the main points in a collection of comments. KPA extracts a set of concise, high-quality sentences or phrases, termed \emph{Key Points}, and maps each of the input sentences to its corresponding key points. The prevalence of each key point is quantified as the number of its matching sentences. KPA summaries are more expressive than the commonly-used word clouds and key phrases, while adding an important quantitative dimension that is missing from plain text summaries. 

The KPA algorithm aims to extract a set of key points that provide high coverage of the data, while removing redundancies. It employs two supervised models: one for assessing the quality of key point candidates, and another one for computing a match score between a sentence and a candidate key point. \citet{bar-haim-etal-2021-every} adapted KPA to business reviews, by introducing several extensions to the original algorithm. In particular, they integrated sentiment analysis into KPA, creating separate summaries for positive and negative sentences. They also developed a specialized key point quality model for the business reviews domain. 
\paragraph*{Entailment Graphs.}

Most of the prior work on entailment graphs has focused on learning entailment relations between predicates, while satisfying some global constraints such as transitivity \cite{berant-etal-2010-global}, soft transitivity \cite{chen-etal-2022-entailment}, and other types of soft constraints \cite{hosseini-etal-2018-learning}.  \citet{levy-etal-2014-focused} extended the notion of entailment graphs to instantiated predicates.  

Most similar to our Key Point Hierarchies are entailment graphs over text fragments, introduced by \citet{Kotlerman2015TextualEG}. Their motivating scenario was summarizing customer feedback, for which they developed a benchmark dataset. However, the text fragments in this dataset were extracted manually. The approach proposed in the current work, which first finds the most salient points in the data using KPA, and then constructs a hierarchy from the extracted key points, allows fully-automatic generation of structured summaries for large collections of opinions, views or arguments. Constructing hierarchies over automatically-extracted key points, which are often noisy and imperfect, represents a more realistic scenario, and makes both manual annotation of KPHs and their automatic construction more challenging.

\section{Key Point Hierarchies}
\label{sec:task}
Figure~\ref{fig:kph_example} illustrates the transformation of a flat key point list into a Key Point Hierarchy (KPH). Formally, given a list of key points $\mathcal{K} = \{k_1, k_2, ..., k_n\}$, we define a KPH $H = (\mathcal{V},\mathcal{E})$ as a directed forest, that is, $H$ is a Directed Acyclic Graph (DAG) where each node has no more than one parent. The vertices $\mathcal{V}$ are clusters of key points $\{C_1, ..., C_m\}$ that convey similar ideas, and the directed edges $\epsilon_{ij} \in \mathcal{E}$ represent hierarchical relations between clusters $C_i$ and $C_j$. Similar to~\citet{Kotlerman2015TextualEG}, a directed edge $C_i \xrightarrow{} C_j$ indicates that the key points in $C_i$ provide elaboration and support for the key points in $C_j$. By transitivity, this relation extends to any two clusters $C_i$ and $C_k$ such that there is a directed path in $H$ from $C_i$ to $C_k$, which we denote as $ C_i \leadsto C_k$. Accordingly, we define $\mathcal{R}(H)$ as the set of directional relations between pairs of \emph{key points} $(x, y)$ that can be derived from $H$ as: 
\begin{equation}
    \mathcal{R}(H) = \{(x, y) ) \ | \ C_x = C_y \lor C_x \leadsto C_y \}
     \label{eq:relations}
\end{equation}
where $C_x, C_y \in \mathcal{V}$ are the clusters of $x$ and $y$ respectively. Considering the example in Figure~\ref{fig:kph_example}, $\mathcal{R}(H)$ includes the relations \example{Housekeeping was fantastic} $\xrightarrow{}$ \example{The personnel were great}, \example{Housekeeping was fantastic} $\xrightarrow{}$ \example{Friendly service all around}, \example{Housekeeping was fantastic} $\xrightarrow{}$ \example{The cleaning crew is great}, and so on.  

We chose a hierarchical representation over a more general graph structure since it results in a simpler output that is easier to consume. In addition, this greatly simplified the annotation process. We found that hierarchical representation works well in practice, as the vast majority of the nodes in our dataset did not have more than one potential parent. This is in line with  previous work, which suggested that entailment graphs tend to have a tree-like structure \citep{berant-etal-2012-efficient}.
\section{\data: A Dataset for Key Point Hierarchies}
\label{sec:dataset}
In this section we present \data{}, a benchmark dataset of key point hierarchies. To build \data{}, we first apply Key Point Analysis to reviews of businesses and products from multiple domains~(§\ref{subsec:kpa}). A KPH is then constructed manually from the set of key points extracted for each business or product~(§\ref{subsec:annotation}). We provide statistics on the resulting dataset, as well as qualitative analysis of the types of inferences it includes~(§\ref{subsec:data_prop}).

\subsection{Key Point Set Generation}
\label{subsec:kpa}

The first step in creating the dataset was to run KPA on the reviews of selected businesses and products. Our implementation follows \citep{bar-haim-etal-2021-every}, who suggested several extensions of KPA for analyzing business reviews.\footnote{Specifically, our implementation follows their \textit{RKPA-FT} configuration, except that we extract the key points for each business independently, and allow each sentence to match multiple key points.} For each business, two separate summaries of positive and negative key points are created. 

To obtain a diverse dataset, we considered three different domains, from two data sources:

\paragraph*{Yelp.} This dataset includes 7M written business reviews, where each business may be classified into multiple categories, in varying levels of granularity. 
    We apply KPA to a sample of businesses that include at least one of the following categories: \textsc{Restaurants}, \textsc{Hotels},  and \textsc{Art \& Entertainment}, and had at least 1,000 reviews. For the KPH annotation, we selected four restaurants (which we refer to as the \textsc{Restaurants} domain), and four businesses categorized as  \textsc{Art \& Entertainment}, out of which three were hotels (hereafter, the \emph{Hotels \& Entertainment} domain, or \textsc{Hotels} for brevity). Each domain includes two positive and two negative KPA summaries. 
    
\paragraph*{Amazon\footnote{\url{https://s3.amazonaws.com/amazon-reviews-pds/tsv/index.txt}}.} This dataset includes over 130M customer reviews for a huge collection of products in \url{Amazon.com} across a wide variety of domains. Here, we focused on laptops and tablets from the \textsc{PC} domain, for which we could expect  a rich and diverse set of key points discussing various aspects such as size, ease of use, design etc. Eventually, we annotated a KPH for three positive and one negative KPA summaries.

\subsection{KPH Annotation}
\label{subsec:annotation}
Annotating complex structures such as KPHs is a challenging task, since it involves global, interdependent decisions. Furthermore, the annotator needs to consider different types of hierarchical relations that may hold between the key points, as we further discuss in Section~\ref{subsec:data_prop}. Finally, user reviews make extensive use of informal and figurative language. For example, \example{The food is outrageous!} should be interpreted as great food; \example{Elevators should go up and down, not diagonal} means that the elevators were scary and \example{Internet was a joke to get to work} indicates a poor WiFi signal.

To overcome these challenges and obtain a high-quality dataset, three annotators individually constructed a KPH for each KPA summary (§\ref{subsubsec:annotation}); The annotators then met to resolve their disagreements and reach a consolidated KPH (§\ref{subsubsec:consolidation}).

\subsubsection{Creating an Initial KPH}
\label{subsubsec:annotation}

To construct an initial KPH, annotators were shown the key points one by one in a descending order according to the number of their matched sentences. For each key point, they first decided whether it conveys the same idea as any previously seen key point, in which case it was added to an existing cluster. If not, a new node was added to the KPH, and the annotator dragged it to its right position in the hierarchy. Since key points with many matches tend to be more general, the key point ordering facilitated top-down construction of the KPH. At any point in the annotation process, annotators had a complete view of the KPH constructed so far, and could adjust it by modifying previous decisions, including both clustering and hierarchical relations. Each KPH was annotated separately by three of the authors and took about one hour to complete per annotator. Our annotation guidelines are detailed in Appendix~\ref{app:guidelines}.

Since the key points were extracted automatically, some of them did not satisfy the desired properties of a key point - a concise and self contained sentence or phrase that discusses a single point with a certain polarity~\citep{bar-haim-etal-2021-every}. To avoid noise in \data{}, annotators could mark such bad key points as candidates for removal from the final KPH. 

As our annotation tool, we used CoRefi~\citep{bornstein-etal-2020-corefi}, an interface for cross-document coreference annotation with \citet{cattan2021scico}'s extension for annotating a forest of clusters, which we adapted to handle key points (see Appendix~\ref{app:annotation}).

\subsubsection{KPH Consolidation}
\label{subsubsec:consolidation}
To obtain the final KPHs, the three annotators met to discuss and resolve the differences in their individual KPHs annotations. This is a complex process because both clusters and the relations between them can differ. We therefore separated the consolidation process into two subsequent stages: clustering and hierarchy.

In the first phase, following the reviewer mode in CoRefi~\citep{bornstein-etal-2020-corefi}, annotators were shown one key point at a time with their original clustering decisions. In case of disagreement, the annotators discussed and reached a joint decision, which automatically modified their original KPH accordingly. At the end of this stage, the initial KPH of each of the annotators was modified to include the exact same nodes. In the second phase, since each key point has a single parent, we could easily identify the remaining disagreements by comparing the parent of each node across the different annotators. To support this consolidation phase, we enhanced CoRefi with the ability to identify and highlight both clustering and hierarchy disagreements between any number of annotators (see Appendix~\ref{app:consolidation} for more details). 

Consolidating multiple annotations was also efficient due to the hierarchical structure of the KPH and took about an hour per KPH. 

\subsubsection{Dataset Quality Assessment} 
\label{subsubsec:quality}

To verify the quality of the resulting dataset, we asked two additional annotators to annotate and consolidate a portion of \data{} (3 \textsc{Restaurants}, 2 \textsc{Hotel} and 2 \textsc{PC}).\footnote{See Appendix~\ref{app:training} for more details about annotators training.} We then evaluated their individual and consolidated KPHs against our consolidated annotation, as follows. In each domain, we compared the two sets of annotated KPHs by taking the union of the KP relations induced by the KPHs in each set (Eq.~\ref{eq:relations}), and computing the F1 score over the two resulting sets of relations. The final F1 was obtained by macro-averaging over the three domains. 

The annotators' performance after consolidation reached an F1 of 0.756, indicating substantial agreement.\footnote{We do not report Kappa because decisions are mutually dependent.} Furthermore, consolidation was shown to increase individual performance by 5-6 points.%

\subsection{Dataset Properties}
\label{subsec:data_prop}

\begin{table}[t]
    \centering
    \resizebox{0.48\textwidth}{!}{
    \begin{tabular}{llcccc}
    \toprule
         && \textsc{Rest} & \textsc{Hotel} & \textsc{PC} & Total\\
         \midrule 
        \#KPHs && 4 & 4 & 4 & 12 \\ 
        \#Key points && 181 & 208 & 128 & 517\\
        \#Filtered KPs && 21 & 17 & 48 & 86\\
        \# $\mathcal{R}(H)$ && 850 & 302 & 266 & 1,418\\ 
        \bottomrule
    \end{tabular}}
    \caption{Statistics of \data{}. $\mathcal{R}(H)$ is the set of key point relations that can be derived from a KPH $H$ (§\ref{sec:task}).}
    \label{tab:stats}
\end{table}

Table~\ref{tab:stats} shows some statistics for the \data{} dataset. Overall, \data{} includes 12 KPHs, 517 key points, and 1,418 key points relations ($\mathcal{R}(H)$) out of the total 24,430 key point pairs. 
Due to its size,  we did not split \data{} into development and test sets, but rather used the entire dataset for evaluation. As described in Section~\ref{subsubsec:annotation}, during the annotation, we filter a relatively small number of key points (14\%), mostly from the \textsc{PC} domain. This is mainly because the key point quality model that we used was not trained on this domain. 

From a qualitative perspective, \data{} has several appealing properties that make it a valuable benchmark for NLI. First, recall that the KPA algorithm aims to remove similar key points to avoid redundancy in the summary~\citep{bar-haim-etal-2020-quantitative}. Hence, remaining equivalent key points in \data{} are mostly non-trivial paraphrases that are challenging to detect (e.g., \example{Took forever to get our room} $\leftrightarrow$ \example{Lines to check in are ridiculous}). In addition, hierarchical relations between key points represent diverse types of inferences. Table~\ref{tab:think_examples} shows a few examples of common relations we observed by analyzing a sample from the dataset. 
Finally, \data{} comprises naturally-occurring texts and relations, coming from real-world data.

\begin{table*}[t]
    \centering
    \resizebox{\textwidth}{!}{
    \begin{tabular}{>{\raggedright}p{4cm}p{14cm}}
    \toprule
    \textbf{Relation Type}  & \textbf{Examples} \\
    \midrule
    \multirow{5}{*}{Support / Elaboration}     & Housekeeping needs worked on $\xleftarrow{}$  The beds weren't even made right \\
    & The room was poorly maintained $\xleftarrow{}$ The air conditioning was not functioning right. \\ 
    & The device itself is so difficult to use $\xleftarrow{}$  Transferring data was a nightmare! \\
    & Customer service is a joke $\xleftarrow{}$  No help moving rooms \\
    \midrule
      \multirow{2}{*}{Part-of} 
      & The \textit{hardware} is fantastic $\xleftarrow{}$ \textit{Sound} is surprisingly good \\
      & The \textit{theatre} is great $\xleftarrow{}$ The \textit{entrance} is absolutely beautiful. \\
    \midrule
    \multirow{2}{*}{IS-A} & The \textit{toiletries} they offer are the worst $\xleftarrow{}$ not even good \textit{shampoo} in room \\
    & \textit{Food} varieties was very limited $\xleftarrow{}$  \textit{Desert} selection was below average as well\\ 
    
      \bottomrule
    \end{tabular}}
    \caption{Examples of relations between key points in \data{}.}
    \label{tab:think_examples}
\end{table*}

\section{Automatic KPH Construction}
\label{sec:method}
We use a two-step approach to automatically build a KPH from a set of key points. In the first step, we predict directional scores between all pairs of key points (§\ref{subsec:local}). In the second step, we  construct a hierarchy based on the local scores~(§\ref{subsec:global}).

\subsection{Scoring Pairwise Key Point Relations}
\label{subsec:local}
Given a pair of key points $(i, j)$, we aim to predict whether a directional relation $i \xrightarrow{} j$ holds between $i$ and $j$, by computing a likelihood score $s(i, j) \in [0, 1]$. We experimented with both existing baselines and new methods we developed for this task. Due to the size of \data, it was not used to fine-tune the scoring models (§\ref{subsec:data_prop}).

\paragraph*{Baselines.} Identifying directional relations between two key points is closely related to two existing tasks: Textual Entailment, also known as Natural Language Inference (NLI) \citep{Dagan2007ThePR} and matching arguments to key points~\citep{bar-haim-etal-2020-arguments}. Accordingly, we implemented two baselines: (1) \textbf{\emph{NLI}}, a RoBERTa model~\citep{Liu2019RoBERTaAR} fine-tuned on the MNLI dataset~\citep{williams-etal-2018-broad} to predict whether $i$ \textit{entails} $j$\footnote{\url{https://huggingface.co/roberta-large-mnli}} and (2) \textbf{\emph{KPA-Match}}, a RoBERTa model trained on the \textit{ArgKP} dataset~\citep{bar-haim-etal-2020-arguments} to predict whether $i$ \textit{matches} $j$, following \citep{bar-haim-etal-2021-every}'s implementation.

\paragraph*{Directional Distributional Similarity.} 
\citet{geffet-dagan-2005-distributional} introduced the distributional inclusion hypothesis for lexical entailment~\citep{geffet-dagan-2004-feature}, which suggests that the context surrounding an entailing word $w_1$ is naturally expected to occur also with the entailed word $w_2$. Specifically, for each word $w$, they built a sparse feature vector where the value of the i-th entry is the PMI of the i-th word in the dictionary with $w$. Many distributional similarity metrics have been proposed to predict directional relations such as hyponymy between a pair of words, based on their distributional feature vectors. Among these methods are WeedsPrec~\citep{weeds-weir-2003-general}, BInc~\citep{szpektor-dagan-2008-learning}, ClarkeDE~\citep{clarke-2009-context} and APinc~\citep{kotlerman-etal-2009-directional}.

In this work, we argue that this distributional inclusion hypothesis may be extended to identify directional relations between two key points. Indeed, if $i \xrightarrow{} j$, it is likely that an input sentence that matches the key point $i$ will also match $j$. For example, the sentence \example{The beds were really comfortable, I literally knocked out as soon as my head touched the pillow.} matches both \example{The beds were awesome} and \example{The rooms are comfortable}. Therefore, we construct a feature vector for each key point $k$, whose length is equal to the number of input sentences. The value at the i-th position in this vector is the likelihood that the i-th sentence matches $k$, as predicted by the KPA matching model~(§\ref{subsec:kpa}). Then, we apply the aforementioned distributional similarity metrics to predict a directional score $s(i, j)$. We only report the performance of \textbf{\emph{APinc}} as it slightly outperformed other metrics. Additionally, we implemented a simple variant of WeedsPrec, in which the entries in the feature vectors are binary (match/no match).  This metric, termed \emph{Binary Inclusion} (\textbf{\emph{BinInc}}), computes the ratio between the number of sentences matched by KPA to both $i$ and $j$ and the number sentences matched to $i$. Intuitively, when most of the sentences that were mapped to $i$ were also mapped to $j$, it is a strong indication that $i \xrightarrow{} j$.

\paragraph*{Combining NLI with Distributional Methods.}
As further discussed in Section~\ref{sec:experiments}, we empirically found that the NLI model and the distributional methods have complementary strengths. The NLI model performs better on \textsc{Restaurants}, whereas the distributional methods perform better on the \textsc{Hotel} and \textsc{PC} domains. Furthermore, even within each domain, those two methods produce very different rankings, as indicated by a low Spearman correlation between their output scores (see Appendix~\ref{app:analysis} for more details).

To take advantage of the strengths of both approaches,  we explored two alternatives for combining BinInc, the best-performing distributional method (as shown in Section~\ref{sec:experiments}), with NLI:
\begin{enumerate}
    \item Averaging the output scores of NLI and BinInc (denoted \textbf{\emph{NLI+BinInc-Avg}}).
    \item Fine-tuning the NLI model on weak labels created by the BinInc model (denoted \textbf{\emph{NLI+BinInc-WL}}). Specifically, we first apply the \textit{BinInc} method to a large number of unlabeled KPA summaries and obtain local scores between all pairs of key points. We then convert these pairwise scores to the NLI format, where we consider all pairs above some threshold as entailment and the others as neutral. Finally, we fine-tune the NLI model on this automatically-generated training data and use the resulting model to predict the local scores $s(i, j)$ on \data. Implementation details and statistics on the silver data are detailed in Appendix~\ref{app:implementation}.
\end{enumerate}

\subsection{Hierarchy Construction}
\label{subsec:global}
We proceed to construct a KPH by determining its semantic clusters and the hierarchical relations between them. Intuitively, we would like to generate a KPH such that the set of pairwise key point relations induced by its structure are consistent with the local directional scores: high-scoring relations should be included, and low-scoring relations should be excluded. We explored several alternatives for constructing a KPH, described below. Each of these methods employs a decision threshold  $\tau$ over the local scores, which needs to be tuned over some development data. 

\paragraph*{Reduced Forest.}
\label{subsubsec:reduced}
\citet{berant-etal-2012-efficient} described a simple transformation of a directed graph $G$ into a forest of clusters. In our case, we start with a graph that includes the key points as nodes, and the directional edges $e(i,j)$ for pairs with local score $s(i,j) > \tau$.%

The reduced forest is constructed as follows: (a) the condensation of $G$ is computed by contracting each strongly connected component into a single vertex that represents a cluster of nodes in $G$. The resulting DAG is transformed into a forest by (b) taking its transitive reduction, and (c) heuristically selecting a single parent for each node with multiple parents. We select the larger cluster as a parent, and as a tie breaker, we use the mean over all the pairwise scores $s(i,j)$ such that $i$ is in the child cluster and $j$ is in the parent cluster. 

As defined by \citeauthor{berant-etal-2012-efficient}, $G$ is a \emph{Forest Reducible Graph (FRG)} if after applying step $b$ above, none of the nodes has multiple parents.

\paragraph*{Tree Node and Component Fix (TNCF).}
\label{subsubsec:berant}
Given a directed graph with local edge weights that are either positive (predicting pairwise entailment between connected nodes) or negative (predicting non-entailment), the optimal entailment graph may be defined as the transitive subgraph in which the sum of the edge weights is maximized \citep{berant-etal-2012-efficient}. \citeauthor{berant-etal-2012-efficient} showed that this problem is NP-Hard, even when further constraining the resulting graph to be forest-reducible. 

To address the computational complexity of finding an exact solution, \citeauthor{berant-etal-2012-efficient} presented an efficient approximation algorithm, termed \emph{Tree-node-fix (TNF)} that generates forest-reducible entailment graphs, and showed empirically that the quality of the resulting graphs is close to the exact solution found via Integer Linear Programming (ILP). Starting from some initial FRG, their algorithm iteratively improves the graph objective function by removing and reattaching one node at a time, while keeping the graph forest-reducible. %

\citet{berant2015efficient} proposed an extension for this algorithm, termed \emph{Tree-Node-and-Component-Fix (TNCF)}, where in each iteration a whole cluster may be re-attached, in addition to individual nodes. We found this extension beneficial. 

Since a KPH is also a forest of clusters, the TNF and TNCF algorithms are directly applicable to our setting. Following \citet{berant-etal-2012-efficient} we defined the edge weights as $w_{i,j} = s(i,j) - \tau$
so that local scores below the threshold $\tau$ are considered negative. 

One difference between the original TNF implementation and ours is the initialization: while they used \citep{berant-etal-2011-global}'s exact solution, computed via ILP for a sparse configuration, we take a simpler approach and start with the reduced forest described above, constructed with the same threshold $\tau$.

\paragraph*{Greedy.}
As an alternative to the TNF/TNCF algorithms, we also adapted the greedy algorithm proposed by \citet{cattan2021scico} for the task of hierarchical cross-document coreference resolution, which also generates a forest of clusters. First, key point clusters are obtained by agglomerative clustering with average linkage and distance threshold of $1-\tau$, where the distance metric between two key points $i$ and $j$ is defined as $1 - min(s(i,j), s(j,i))$. 

Second, we define the score of the directional edge between two clusters $(\mathcal{C}_1, \mathcal{C}_2)$ as the average of the $s(i,j)$ scores between the key points in the two clusters:
\begin{equation}
\label{eq:clusters_scores}
 S(\mathcal{C}_1, \mathcal{C}_2) = \frac{1}{|\mathcal{C}_1|\cdot|\mathcal{C}_2|} \sum_{i \in \mathcal{C}_1} \sum_{j \in \mathcal{C}_2} s(i, j)   
\end{equation}
The KPH is constructed by repeatedly adding the highest-scoring edge (if the score is above the $\tau$ threshold), skipping edges that would violate the definition of the KPH as a directed forest. The process is terminated when no more edges can be added. 

Note that unlike the TNF/TNCF algorithms, the Greedy algorithm does not modify existing clusters and edges in each iteration, but only adds new edges. 
\paragraph*{Greedy with Global Score (Greedy GS).}
One limitation of the Greedy algorithm is that the edge scoring function is \emph{local} and hence ignores indirect relations between clusters that would result from adding the edge. For example, consider a KPH with three clusters $\{A, B, C\}$ such that $B \xrightarrow{} A$. The criterion to add the edge $C \xrightarrow{} B$ will consider only $S(C, B)$ but not $S(C, A)$, which corresponds to the indirect relation $C \leadsto A$. 
To address this issue, we modified the algorithm to consider the relations between each cluster and all its ancestors in the resulting KPH, as follows:
\begin{align}
    E_{k+1} &= E_k \cup \argmax_{\epsilon \in E^\ast \setminus E_{k}} {O(\mathcal{V}, E_k \cup \epsilon)} \label{eq1} \\
    \vspace{-1pt}
    O(\mathcal{V}, \mathcal{E}) &= \sum_{\mathcal{C}_i \in \mathcal{V}} \sum_{\mathcal{C}_j \in A_{\mathcal{V}, \mathcal{E}}(\mathcal{C}_i)} S(\mathcal{C}_i, \mathcal{C}_j) \label{eq2}
\end{align}
where $E_k$ is the set of edges in the resulting KPH after $k$ iterations, $E^\ast$ is the set of all edges scoring above $\tau$ and $A_{\mathcal{V}, \mathcal{E}}(C)$ denotes the set of ancestors of $C$ in  $H(\mathcal{V}, \mathcal{E})$.

\section{Evaluation}
\label{sec:experiments}

\paragraph*{Predicting Local Pairwise Relations.}

\begin{figure*}[t]
    \centering
    \includegraphics[width=\textwidth]{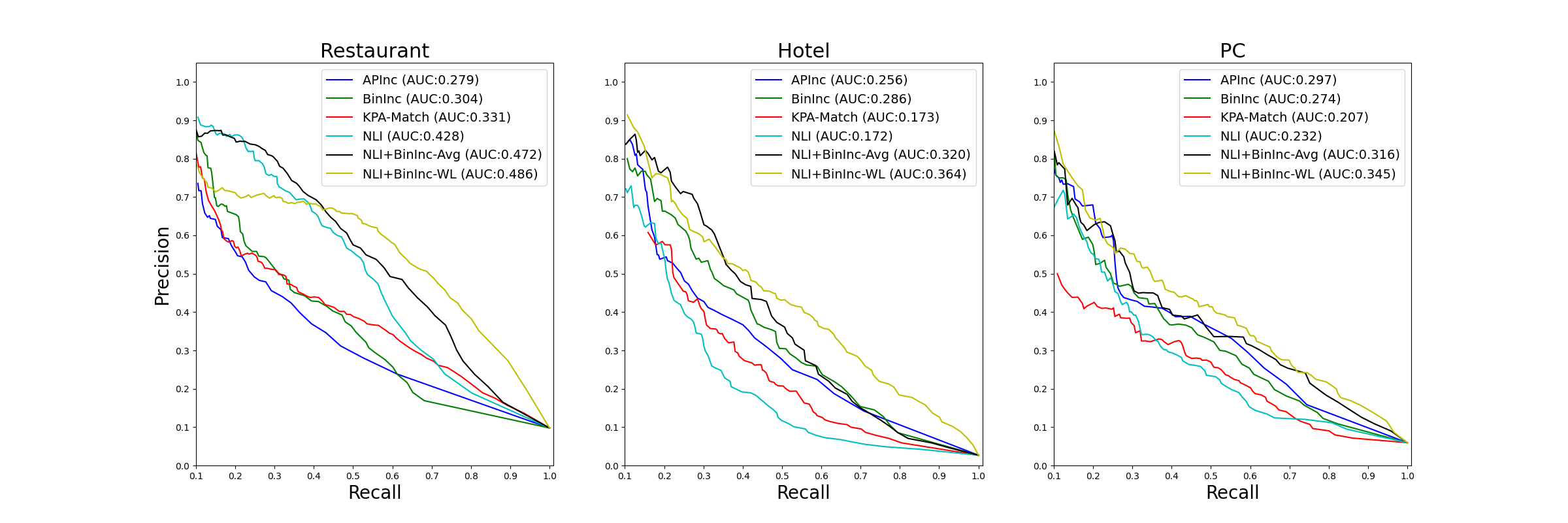}
    \caption{Precision-Recall curves of local scoring methods on \textsc{Restaurant}, \textsc{Hotel} and \textsc{PC}.}
    \label{fig:local_results}
\end{figure*}
Figure~\ref{fig:local_results} compares the performance of the different local scoring methods (§\ref{subsec:local}). For each domain, we consider all the key point pairs in the dataset, and show the Precision/Recall curve and the Area Under the Curve (AUC) for each method. AUC results are also summarized in Table~\ref{tab:all_results}.  

We first observe that applying the \emph{KPA-match} model indirectly via the distributional methods (\emph{APinc} and \emph{BinInc}) outperforms its direct application in two out of the three domains, and increases the average AUC from 0.237 to 0.277/0.288, respectively. The \emph{NLI} model has a clear advantage over the distributional methods in the \textsc{Restaurants} domain, but is much worse for \textsc{Hotel} and \textsc{PC}. Both \emph{NLI+BinInc-Avg} and \emph{NLI+BinInc-WL} models are able to combine the complementary strengths of \emph{NLI} and \emph{BinInc} and outperform all the stand-alone models. Model combination via weak labeling (\emph{NLI+BinInc-WL}) achieves the best performance in all three domains by a large margin (+0.11 average AUC improvement over the best stand-alone method). 

To further assess the contribution of model combination in the weak labeling setting, we also tested a configuration in which the silver data is labeled by the NLI model (denoted \textbf{\emph{NLI-WL}}). The results are shown on the last row of Table~\ref{tab:all_results}. While the performance is better than NLI alone (demonstrating the value of weak labeling), it is still far below \emph{NLI+BinInc-WL}. Overalll, the results affirm the importance of both model combination and the weakly-labeled data for local scoring performance. 

\paragraph*{Hierarchy Construction.} Next, we compare different methods for constructing a KPH from the set of local pairwise scores  (§\ref{subsec:global}). We use the scores from the best performing local method, \emph{NLI+BinInc-WL}, as found in the previous experiment.

We use the F1 measure as defined in Section~\ref{subsubsec:quality} as our evaluation metric, similar to \citet{Kotlerman2015TextualEG}. Since \data{} has no development set (§\ref{subsec:data_prop}), we employ a leave-one-out scheme to tune the threshold $\tau$. Specifically, for each KPA summary $S$, we find the threshold that maximizes the F1 score of the three other KPHs in the same domain and predict a KPH for $S$ using this threshold. We then compute the F1 score for the predicted KPHs in each domain. 

The results are summarized in Table~\ref{tab:global}. \emph{TNCF} achieves the best overall performance on \data{} with an average F1 of 0.526, substantially improving the \emph{Reduced Forest} baseline. The \emph{Greedy GS} algorithm is the top performer in the Restaurants domain (F1=0.641).  Adding a global scoring function to the greedy algorithm improves the performance by 0.059 (from 0.45 to 0.509). 

 We also evaluated the quality of the predicted relations using only the local scores, with a threshold determined via leave-one-out, as before (last row in Table~\ref{tab:global}). While the resulting set of relations may not represent a valid hierarchy, it still provides an interesting reference point for comparison with the various KPH construction algorithms. We can see that both \emph{Greedy GS} and \emph{TNCF} improve the local results by a substantial margin (+0.028 and +0.045, resp.). These two global methods not only satisfy the constraints of generating a valid KPH, but also improve the pairwise relation prediction of the local scorer.  
\begin{table}[t]
    \centering
    \resizebox{0.48\textwidth}{!}{
    \begin{tabular}{llcccc}
    \toprule
        && \textsc{Rest} & \textsc{Hotel} & \textsc{PC} & Avg. \\
    \midrule
    NLI && 0.428	& 0.172	& 0.232	& 0.277\\ 
    KPA-Match &&  0.331 &	0.173 &	0.207	& 0.237\\
    APinc && 0.279 &	0.256 &	0.297	& 0.277\\ 
    BinInc && 0.304 & 0.286 & 0.274	& 0.288\\ 
    \midrule
    NLI+BinInc-Avg && 0.472	& 0.320	& 0.316	& 0.369\\
    NLI+BinInc-WL && \textbf{0.486}	& \textbf{0.364}	& \textbf{0.345}	& \textbf{0.398}\\
    \midrule
     NLI-WL && 0.466	& 0.243	& 0.233	& 0.314\\
    \bottomrule
    \end{tabular}}
    \caption{Evaluation of local scoring methods (AUC for Recall $\ge$ 0.1) }
    \label{tab:all_results}
\end{table}
\begin{table}[t]
    \centering
    \resizebox{0.48\textwidth}{!}{
    \begin{tabular}{llcccc}
    \toprule
        && \textsc{Rest} & \textsc{Hotel} & \textsc{PC} & Avg. \\ 
    \midrule
    
    Reduced Forest && 0.597	& 0.335	& 0.396	& 0.443\\ 
    
    TNCF && 0.614	&\textbf{ 0.460}	& \textbf{0.505} &	\textbf{0.526} \\ 
    Greedy && 0.512 & 0.424 & 0.416 & 0.450 \\ 
    Greedy GS && \textbf{0.641} & 0.433 & 0.451 & 0.509 \\ 
    \midrule
    Local (no tree) && 0.568	& 0.437	& 0.439	& 0.481\\
    \bottomrule
    \end{tabular}}
    \caption{Evaluation of hierarchy construction algorithms (F1 scores). All methods use the \emph{NLI+BinInc-WL} local scores.}
    \label{tab:global}
\end{table}

\section{Conclusion}
We introduced Key Point Hierarchies as a novel representation for structured, expressive opinion summaries. We explored several approaches for automatic hierarchy construction from a given set of key points, which were evaluated on a new benchmark dataset we developed for this task. We also proposed a novel distributional representation for key points, which we leveraged via weak supervision to achieve substantial improvement on the subtask of predicting pairwise key point relations. While our initial results are promising, there is still much room for improvement, and we hope that releasing our dataset would encourage the community to further promote this line of research.

\section*{Limitations}
Key Point Hierarchies may be valuable for summarizing opinions and views in multiple domains, including reviews, survey responses, customer feedback, political debates etc. However, in this work, we only demonstrated their value for business and product reviews, leaving other types of data to future work. Also, we only attempted to create KPHs for English reviews, for which an abundance of resources is available, including a huge number of written reviews and high-quality trained  models, e.g. for NLI and key point matching. Applying these methods to low-resource languages is expected to be far more challenging. Finally, the quality of the resulting KPHs depends on the quality of the extracted key points provided as input, which may vary across different domains. To alleviate this problem in \data{}, we manually filtered out problematic key points from the dataset (§\ref{subsec:annotation}).

\section*{Acknowledgments}

The first author is partially supported by the PBC fellowship for outstanding PhD candidates in data science.

\bibliography{anthology,custom}
\bibliographystyle{acl_natbib}

\appendix
\section{Data Collection}
\label{sec:data}

\subsection{Annotation Guidelines}
\label{app:guidelines}

We began the annotation process of \data{} by drafting guidelines in which we describe the KPH structure (§\ref{sec:task}) and define the annotation task as follows. \example{Given two key points A and B, (1) if A and B roughly convey the same idea or opinion, they should be clustered together in the same node (e.g. Friendly service all around vs. Staff was nice and helpful) and (2) if B elaborates on A and supports it, then B should be placed under A in the hierarchy (e.g., the rooms are comfortable $\xleftarrow{}$ The bed was very comfy)}. Importantly, as key points are automatically extracted from human reviews written by different people in their own vocabulary, we advise to ignore subtle differences because they do not reflect different opinions. For example, \example{Not much choice of fruits and desserts} and \example{Dessert selection was below average as well} should be considered equivalent because \example{Dessert} usually includes fruits. 

\subsection{Annotation} 
\label{app:annotation}

\begin{figure*}[t]
    \centering
    \includegraphics[width=0.9\textwidth, frame]{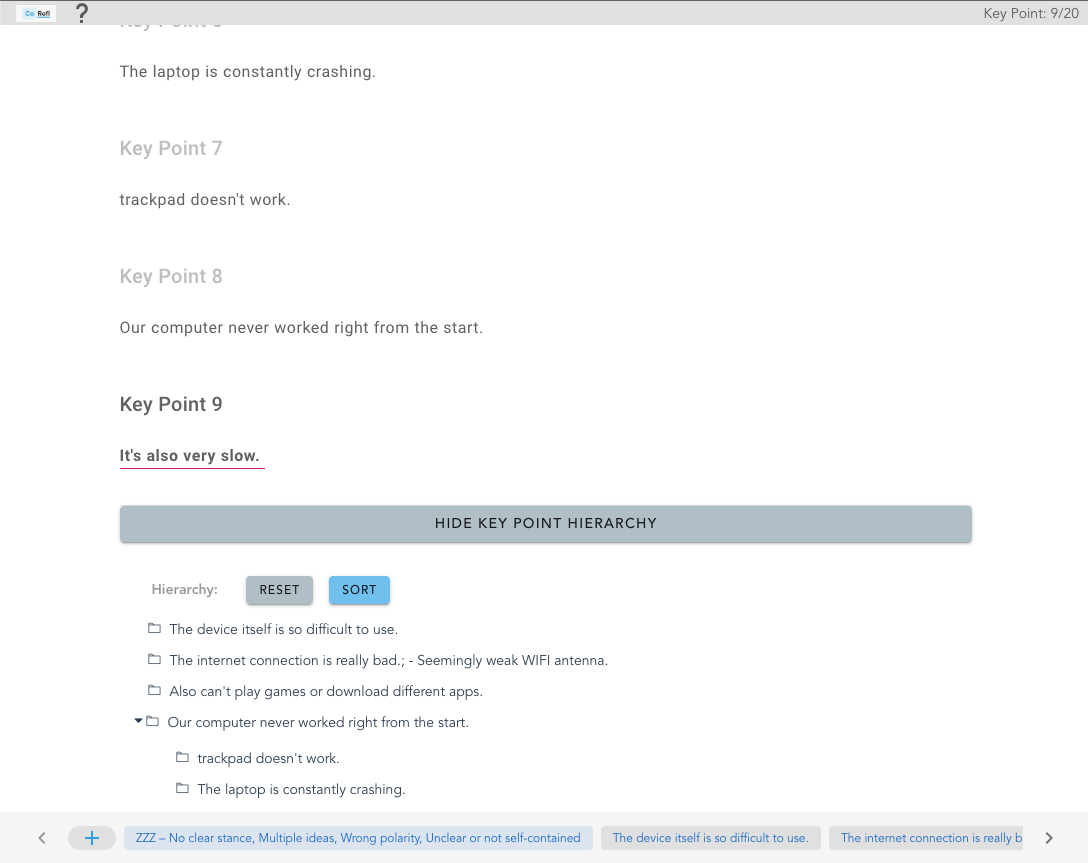}
    \caption{\textsc{CoRefi} annotation interface adapted to annotate \data{}. }
    \label{fig:annotation}
\end{figure*}

Figure~\ref{fig:annotation} shows the \textsc{CoRefi} interface that we use to annotate \data{}. For each key point, annotators decide whether to add it to an existing cluster or to create a new node in the hierarchy.  

\subsection{Consolidation}
\label{app:consolidation}

\begin{figure*}[t]
    
         \centering
         \includegraphics[width=0.9\textwidth, frame]{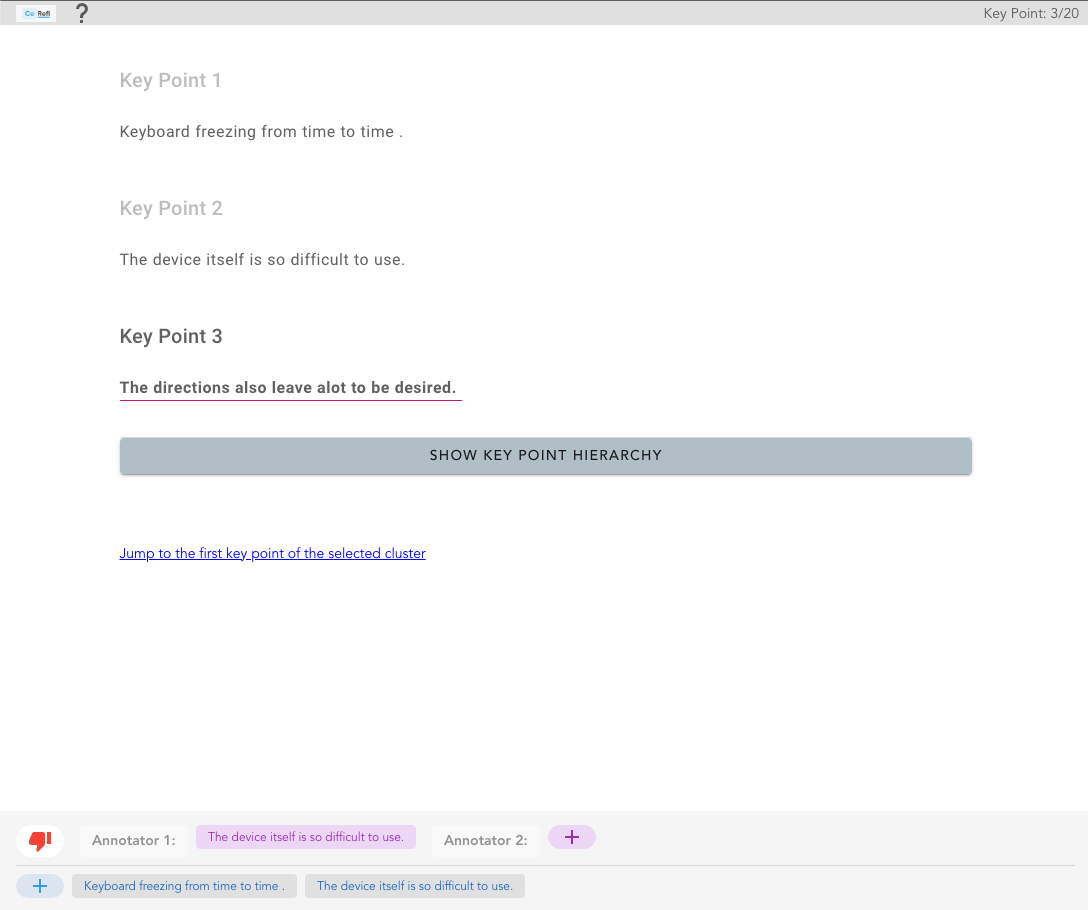}
         \caption{Clustering step. The thumb-down at the bottom left of the screen indicates a clustering disagreement between the annotators for Key Point 3: \example{The directions also leave a lot to be desired}. Annotator A1 assigned it to \example{The device itself is so difficult to use} while annotator A2 created a new cluster, as indicated in purple.}
         \label{fig:cons1}
     \end{figure*}
     \begin{figure*}
         \centering
         \includegraphics[width=0.9\textwidth, frame]{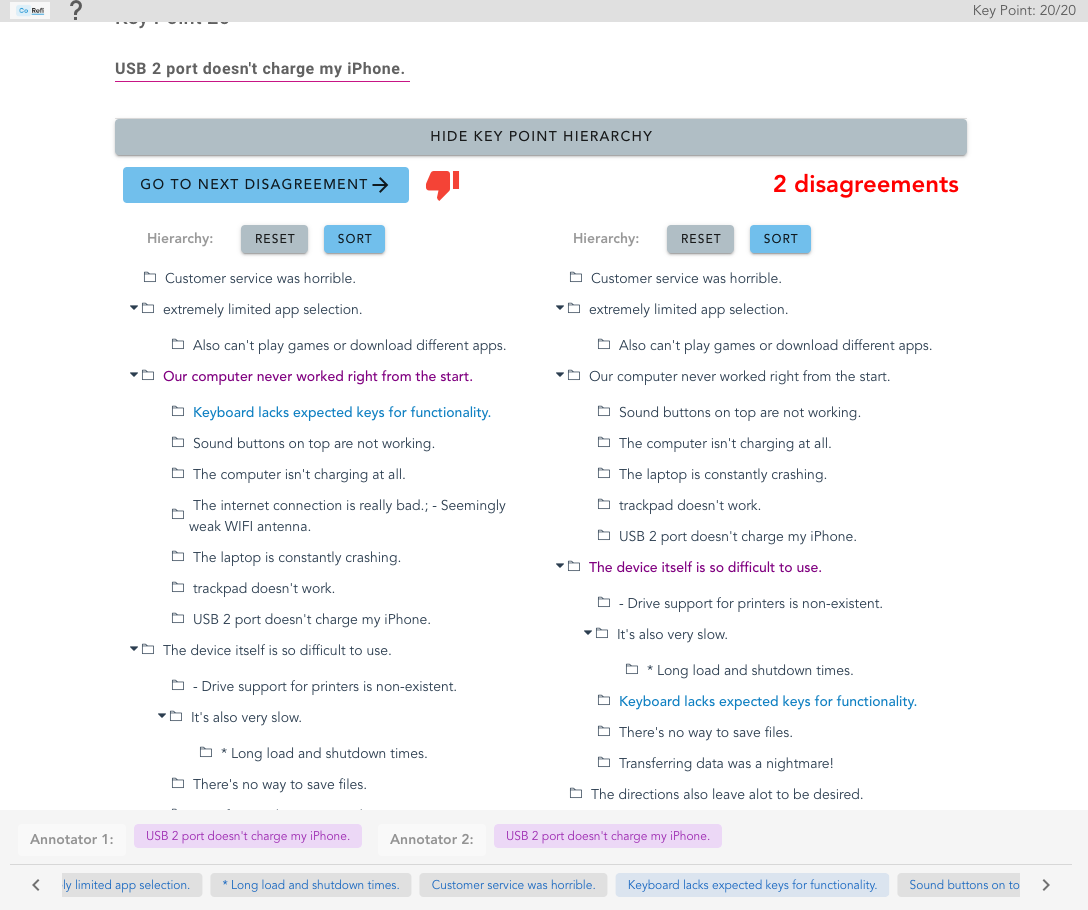}
         \caption{Consolidation of hierarchical relations. The cluster \example{Keyboard lacks expected keys for functionality.} is highlighted in blue in both KPHs because the two annotators placed it under different parents (colored in violet in both KPHs). 
         }
         \label{fig:cons2}

\end{figure*}

As described in the paper~(§\ref{subsubsec:consolidation}), we split the consolidation stage into two subsequent steps: clustering and hierarchy, illustrated in Figures~\ref{fig:cons1} and~\ref{fig:cons2}.

For the clustering step (Figure~\ref{fig:cons1}), we extend the reviewer algorithm in \textsc{CoRefi}~\citep{bornstein-etal-2020-corefi} with the ability to review multiple annotations for the same input. In case of disagreement, we display a red thumb-down at the bottom left of the annotation interface and the annotators discuss to reach a joint decision.

Each clustering decision automatically modifies their original KPHs. Considering the example in Figure~\ref{fig:cons1} with a clustering disagreement for the key point \example{The directions also leave a lot to be desired (KP1)}: annotator A1 grouped it together with \example{The device itself is so difficult to use (KP2)} whereas annotator A2 left it as a standalone node in the KPH (indicated by the + button in purple). Now, if A1 and A2 decide to follow A1's decision, A2's original KPH will be automatically modified to include a grouped node \{\emph{The device itself is so difficult to use, The directions also leave a lot to be desired}\} (instead of two separated nodes) whose children will be the concatenation of the initial children of \emph{KP1} and \emph{KP2}. On the other hand, if A1 and A2 decide to follow A2's decision, a new node \example{The directions also leave a lot to be desired} will be added in A1's KPH. In this case, the children of the initial grouped node will stay under \example{The device itself is so difficult to use}. This automatic process ensures that the original KPHs will include the exact same nodes.

In the second step, as shown in Figure~\ref{fig:cons2}, as the nodes in the two KPHs are identical, a disagreement will occur when a cluster $C \in \mathcal{V}$ has a different direct parent in each KPH. To identify the next disagreement, annotators can click on the ``Go To Next Disagreement'' button to highlight the key point in blue and its direct parent in violet on both KPHs. Once all hierarchical disagreements have been resolved, the structure of both KPHs will be identical and the annotators can submit their consolidated KPH.

\subsection{Annotators Training}
\label{app:training}

To assess the quality of \data{}~(§\ref{subsubsec:quality}), we provided a team of in-house annotators with the same annotation guidelines~(§\ref{app:guidelines}), while explicitly mentioning the purpose of the data collection. Following~\citep{bornstein-etal-2020-corefi}, we also provided them an automated walk-through tutorial to get familiar with the tool functionalities~(§\ref{app:annotation}). As part of the training, we asked the annotators to construct a KPH for 2 different businesses and gave them detailed feedback. Finally, we gave them a test and proceeded with the annotators who passed the test. %

\section{Implementation Details}
\label{app:implementation}

As described in Section~\ref{subsec:local}, our best local scorer is obtained by fine-tuning an NLI model on weakly-labeled data, automatically collected as follows. We first applied KPA to reviews from 152 \textsc{Yelp} businesses. The resulting KPA summaries included 38 key points on average. We then ran the \emph{BinInc} method on all possible key point pairs in each KPA summary. After fixing the decision threshold to 0.5, we obtained 5,379 positive pairs and 295K negative pairs. In the final dataset that was used to train the model, we downsampled the negative examples so that the ratio between positive and negative examples was 1:5.\footnote{We experimented with multiple ratios (1:1, 1:2, 1:3, 1:5, 1:10) as well as considering all the pairs and found that the 1:5 ratio achieves the best performance.}

We train our model using PyTorch~\citep{NEURIPS2019_9015}, PytorchLightning~\citep{falcon2019pytorch} and the Transformers library~\cite{wolf-etal-2020-transformers} for 5 epochs with a batch size of 64 and a learning rate of 1e-7.

\section{Analysis}
\label{app:analysis}

\begin{figure*}[t]
    \centering
    \includegraphics[width=\textwidth]{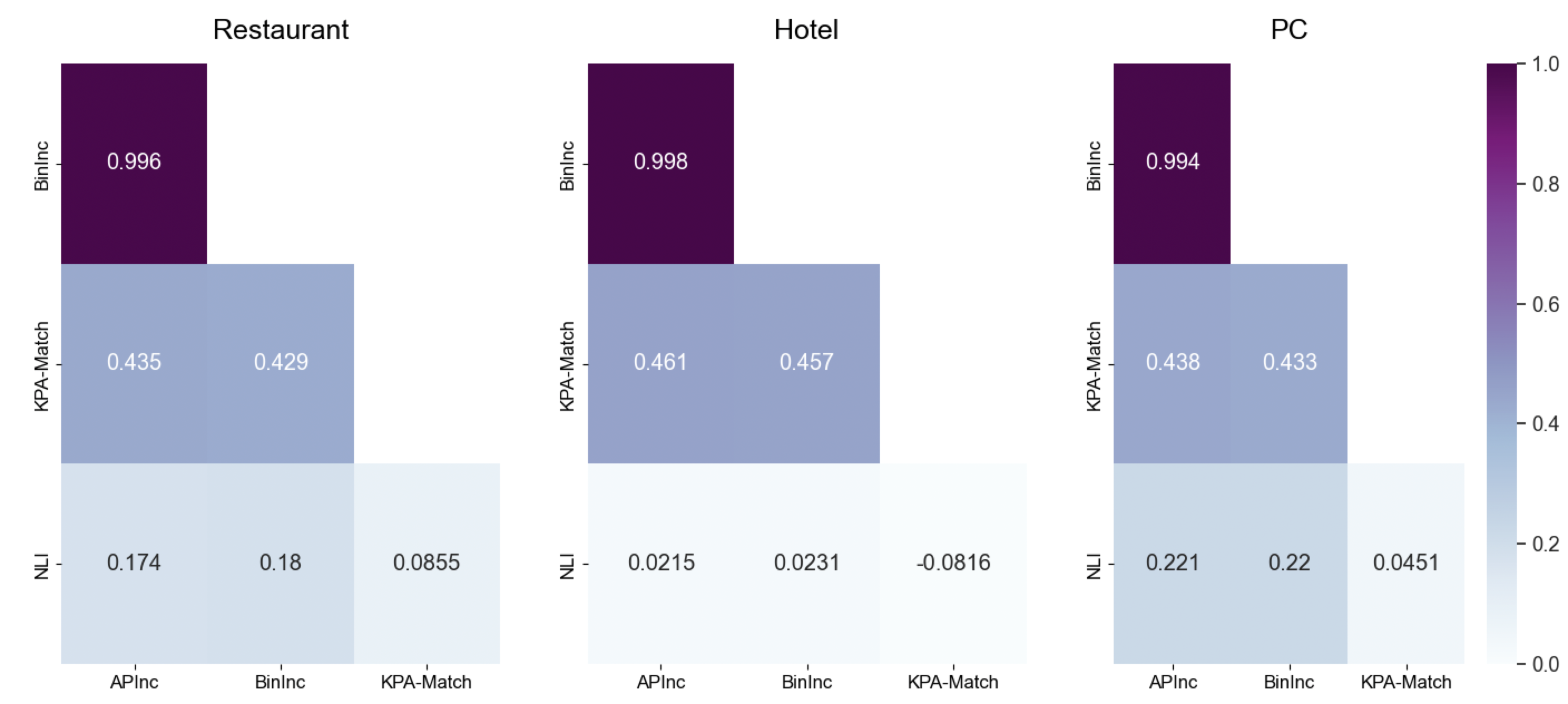}
    \caption{Spearman correlations between the scores of the local methods}
    \label{fig:local_correlations}
\end{figure*}

Figure~\ref{fig:local_correlations} shows the Spearman correlation coefficients between the output scores of the different local methods that we define in Section~\ref{subsec:local}. NLI has a low correlation with the distributional methods (\emph{APinc} and \emph{BinInc}) in each of the three domains. This indicates that NLI and the distributional methods rank the key point pairs quite differently.  

\section{Datasets}
\begin{itemize}
\item The Yelp and Amazon datasets used in this work have been released for academic use, and accordingly, we have only used them for academic research.
\item The authors have reviewed the \data{} dataset and verified that it does not contain any personal information or offensive content. 
\end{itemize}

\end{document}